\documentclass[letterpaper, 10 pt, conference]{ieeeconf}  % Comment this line out if you need a4paper

\IEEEoverridecommandlockouts            
\usepackage{graphics} % for pdf, bitmapped graphics files
\usepackage{epsfig} % for postscript graphics files
\usepackage{mathptmx} % assumes new font selection scheme installed
\usepackage{xcolor}
\usepackage[font=scriptsize]{caption}
\usepackage{subcaption}
\usepackage{times} % assumes new font selection scheme installed
\usepackage{amsmath} % assumes amsmath package installed
\usepackage{amssymb}  % assumes amsmath package installed
\usepackage{sidecap}  %required for side captions
\usepackage{CJKutf8} % to use Chinese characters in text
\usepackage[
    backend=biber,
    style=ieee,
  ]{biblatex}
\usepackage{hyperref}
\hypersetup{
    colorlinks=true,
    linkcolor=blue,
    filecolor=magenta,      
    urlcolor=cyan,
    citecolor=teal,
}
\makeatletter
\newcommand*\rel@kern[1]{\kern#1\dimexpr\macc@kerna}
\newcommand*\widebar[1]{%
  \begingroup
  \def\mathaccent##1##2{%
    \rel@kern{0.8}%
    \overline{\rel@kern{-0.8}\macc@nucleus\rel@kern{0.2}}%
    \rel@kern{-0.2}%
  }%
  \macc@depth\@ne
  \let\math@bgroup\@empty \let\math@egroup\macc@set@skewchar
  \mathsurround\z@ \frozen@everymath{\mathgroup\macc@group\relax}%
  \macc@set@skewchar\relax
  \let\mathaccentV\macc@nested@a
  \macc@nested@a\relax111{#1}%
  \endgroup
}
\makeatother
\title{\LARGE 
\textbf{Technical Report on: Tripedal Dynamic Gaits for a Quadruped Robot}}

\author{Abriana Stewart-Height$^{1,2}$ and Daniel E. Koditschek$^{1,2}$% <-this % stops a space
\thanks{$^{1}$ Department of Electrical \& Systems Engineering, University of Pennsylvania, Philadelphia, USA
}%
\thanks{$^{2}$General Robotics, Automation, Sensing, \& Perception (GRASP) Lab, University of Pennsylvania, Philadelphia, USA {\tt\small (abrianas, kod) @seas.upenn.edu}}%
}

\begin{document}
\maketitle
\thispagestyle{empty}
\pagestyle{empty}

%%%%%%%%%%%%%%%%%%%%%%%%%%%%%%%%%%%%%%%%%%%%%%%%%%%%%%%%%%%%%%%%%%%%%%%%%%%%%%%%
\begin{abstract}
A vast number of applications for legged robots entail tasks in complex, dynamic environments. But these environments put legged robots at high risk for limb damage. This paper presents an empirical study of fault tolerant dynamic gaits designed for a quadrupedal robot suffering from a single, known ``missing'' limb. Preliminary data suggests that the featured gait controller successfully anchors a previously developed planar monopedal hopping template in the three-legged spatial machine. This compositional approach offers a useful and generalizable guide to the development of a wider range of tripedal recovery gaits for damaged quadrupedal machines.
\end{abstract}

%%%%%%%%%%%%%%%%%%%%%%%%%%%%%%%%%%%%%%%%%%%%%%%%%%%%%%%%%%%%%%%%%%%%%%%%%%%%%%%%
\section{INTRODUCTION}
\subsection{Motivation}

For legged robots to assist with essential outdoor activities (such as environmental preservation and restoration, space exploration, search and rescue, wildlife population monitoring, etc.), they must be able to perform desired tasks (e.g. data collection, marking objects, and capturing images) while moving in complex, dynamic environments. Ideally, a robot in its desired state would be able to complete its given assignment and suffer little to no damage while doing so. Yet in the real world, mishaps occur, and in the challenging environments we seek to place our robots, it's almost certain that damage will occur. Legged robots operating in unstructured areas would benefit greatly from the ability to recover from bodily damage. For example, if a legged machine loses a limb while operating in a human restricted environment, a gait developed to adapt to the damage could allow the robot to transport itself to a secure area, where it could be safely retrieved.

While the problem of fault tolerance is actively pursued in robotics \cite{zhang_resilient_2017, zambella_integrated_2020,yueqi_yang_fault-tolerant_2018} and the broader engineering literature \cite{gao_survey_2015,jeong_fault_2019,hu_advanced_2020},
animals offer engineers championship examples of athletic locomotive adaptation to permanent limb damage \cite{goldner_kinematic_2015, wilshin_limping_2018, escalante_rapid_2020}. The ability to develop compensatory gaits in response to unanticipated limb damage is crucial to their survival. Robots that recover effective dynamic behaviors following severe limb damage or loss would greatly benefit their developers and users. Ordinarily, if a legged robot breaks a limb while working in an inaccessible location, it is abandoned as immobile and unreachable, costing the robot, lost time, and, likely, lost data. Equipping robots with fault tolerant gaits could prevent such losses.
\begin{figure}[htp]
    \centering
    \includegraphics[width=0.47\textwidth]{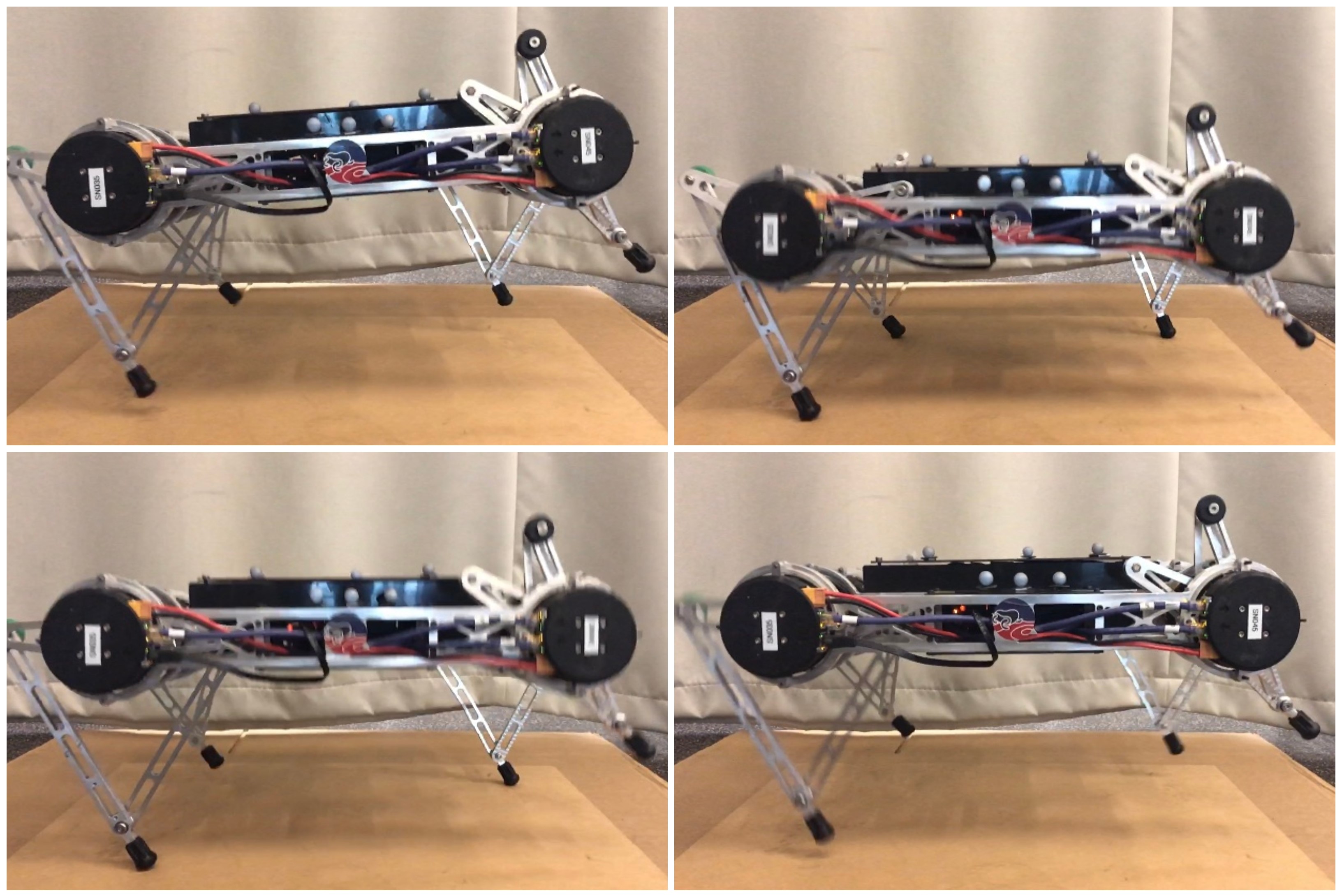}
    \caption{The experimental platform used in this work, Minitaur \cite{kenneally_design_2016}, an underactuated quadrupedal robot, executing vertical tripedal pronking (Sec. \ref{sec:tripronk}).}
    \label{fig:pronk}
\end{figure}
\subsection{Related Work}
The challenge of developing robots that can adapt to changes to their physical form like animals has been explored in past research \cite{cully_robots_2015, goldschmidt_biologically-inspired_2014,zhang_resilient_2017}. Robots described in this literature can complete their desired tasks despite sudden changes to their mechanical structures such as broken appendages, damaged sensors, and limb loss. %This literature describes a variety of machines that exhibit some ability to overcome sudden changes in their environments as well as their mechanical structures and still complete their desired tasks. 
These past adaptations to physical damage have involved a change in: 1) the physical morphology \cite{jakimovski_self-reconfiguring_2009, tolley_resilient_2014,kano_brittle_2017, kriegman_automated_2019, wang_automatic_2019}; 2) the gait generated\footnote{Please note that this list has been condensed. If authors have multiple papers combining the same robot and control strategy, only the most recent paper is mentioned} \cite{johnson_disturbance_2010, chatzilygeroudis_reset-free_2016,ren_multiple_2015,chattunyakit_bio-inspired_2019,manglik_adaptive_2016,miguel-blanco_general_2020,bossens_learning_2020,xu_gait_2020,zhou_research_2020,kon_gait_2020,mailer_evolving_2021,bing_energy-efficient_2020,verma_deep_2020,green_design_2016,park_automated_2013,bongard_resilient_2006,kaushik_fast_2021,song_rapidly_2020,lee_three-legged_2000,zhang_novel_2021,cui_fault-tolerant_2022,gor_fault_2018,chen_fault-tolerant_2014,yang_fault_2012,yang_fault-tolerant_2009,thor_generic_2021,you_bo_instability_2021,spenneberg_stability_2004,vonasek_failure_2015,dujany_emergent_2020,christensen_fault-tolerant_2014,saputra_neural_2020,du_fault_2017,chen_gait_2015,pana_fault-tolerant_2008}; or 3) both the physical structure and gait employed \cite{akhlaq_biologically_2014,tam_fall_2016,akrour_damage_2019,guo_robots_2019}. In this paper, we narrow our scope to join the second body of literature, research that addresses creating alternate gaits for legged robots in response to limb damage.

Presently, the large majority of commercial robots that are intended for application in unstructured outdoor environments (where limb damage is more likely to occur) are quadrupeds \cite{anybotics, boston_dynamics, ghost_robotics}. Moreover, there is a wealth of research in the biomechanical literature on limb damage recovery in biological quadrupeds \cite{fuchs_proprioceptive_2012, hogy_kinematic_2013,goldner_kinematic_2015}. Therefore, we further restrict our scope to quadrupedal machines. 

\subsubsection{Fault detection}
A range of potential consequences arise when a quadruped's limb is damaged such as imbalance, capsizing, and limited or no movement in any direction. Several researchers have approached identifying unanticipated or anticipated failures by first detecting the fault, then selecting a behavior to address it.

Model-based optimization techniques have frequently been used to diagnose unanticipated faults. Past work of this kind  monitors the robot's whole body \cite{bongard_resilient_2006, chen_fault-tolerant_2014, cui_fault-tolerant_2022} or its legs only \cite{gor_fault_2018, miguel-blanco_general_2020, zhang_novel_2021}, and flag a sensed change as a fault. Optimization methods determine the best reaction to the reported fault.

In contrast, past research addressing anticipated damage in legged systems has focused heavily on simulations, which predict all potential failures for a given robot and generate functional gaits for each one. Several authors have proposed evolutionary-based algorithms (EAs) to address anticipated faults \cite{park_automated_2013,chattunyakit_bio-inspired_2019}. EAs often need large search spaces to provide accurate results, making them computationally expensive \cite{park_automated_2013} and time-consuming \cite{zhang_resilient_2017}. The exception to using evolutionary techniques for anticipated fault detection, \cite{kaushik_fast_2021} proposed a meta-learning algorithm that allowed their robot to adapt its model to damage within a few steps. but their method performed poorly with a real robot due to a high reality gap.

\subsubsection{No Fault Detection}
To concentrate only on fault tolerant gaits, some researchers have proposed methods that assume faults are known. Several authors have developed quasi-static gaits for known locked joint failures. \cite{pana_fault-tolerant_2008} derived a periodic gait for straight-line and crab walking for a quadruped robot with a locked joint failure. This type of fault allows the afflicted limb to maintain ground contact but restricts movement.\cite{yang_fault_2012} modeled the constrained motion of a single locked joint with flat shaped feet. \cite{christensen_fault-tolerant_2014} introduced an online learning strategy for their modular robot. \cite{chen_gait_2015} proposed a fault-tolerant gait planning method, based on screw theory. The exception to implementing quasi-static gaits for fault tolerance, \cite{saputra_neural_2020}, proposed a central pattern generated (CPG) model with sensorimotor coordination. The authors implemented their dynamic gait pattern generator on a simulated and real robot to compensate for locked joint failures on different terrains. In contrast, amputated limbs have rarely been addressed in the fault tolerant literature for quadruped robots \cite{lee_three-legged_2000, di_carlo_dynamic_2018}. 

The preponderance of prior literature on quasi-static locked limb failures motivates our interest in dynamical gaits that can address the problem of non-load bearing (e.g., amputated) limbs. The virtue of dynamical locomotion in the context of compensatory locomotion arises, first, from the aim to avoid further body damage incurred by (repeated ground shocks consequent upon) a quasi-static gait performed by a statically unstable three-legged quadruped. Moreover, a longstanding hypothesis in the legged locomotion literature \cite{Hoyt_Taylor_1981} holds that dynamical gaits emerge to reduce the cost of transport as targeted speed is increased. In this context, the pronking gait we focus on here would not normally command priority because it is less efficient than galloping and trotting over its putative range of useful speeds \cite{McMahon_1985}.  However, as intuition (and our empirical results) suggest, the pronking gait seems easiest to stabilize dynamically and hence represents a useful entry into this new problem domain.

\subsection{Contributions and Organization}
This paper addresses empirically the problem of dynamical gait recovery from a known non-load bearing limb, in contrast to the traditional premise of a (load-bearing) locked joint. Specifically, we present a steady-state fore-aft tripedal pronking gait for a power autonomous, unconstrained quadrupedal robot with no external sensing. While tripedal gaits have been previously reported for machines explicitly designed with three legs \cite{feller_mechanical_2022,heaston_strider_2007,karimi_walk--roller_2012}, to the best of our knowledge, this is the first time a tripedal dynamic gait has been implemented on a physical quadrupedal robot.
%
% made the figure larger
\begin{figure}[htbp]
    \centering
    \includegraphics[width=0.47\textwidth]{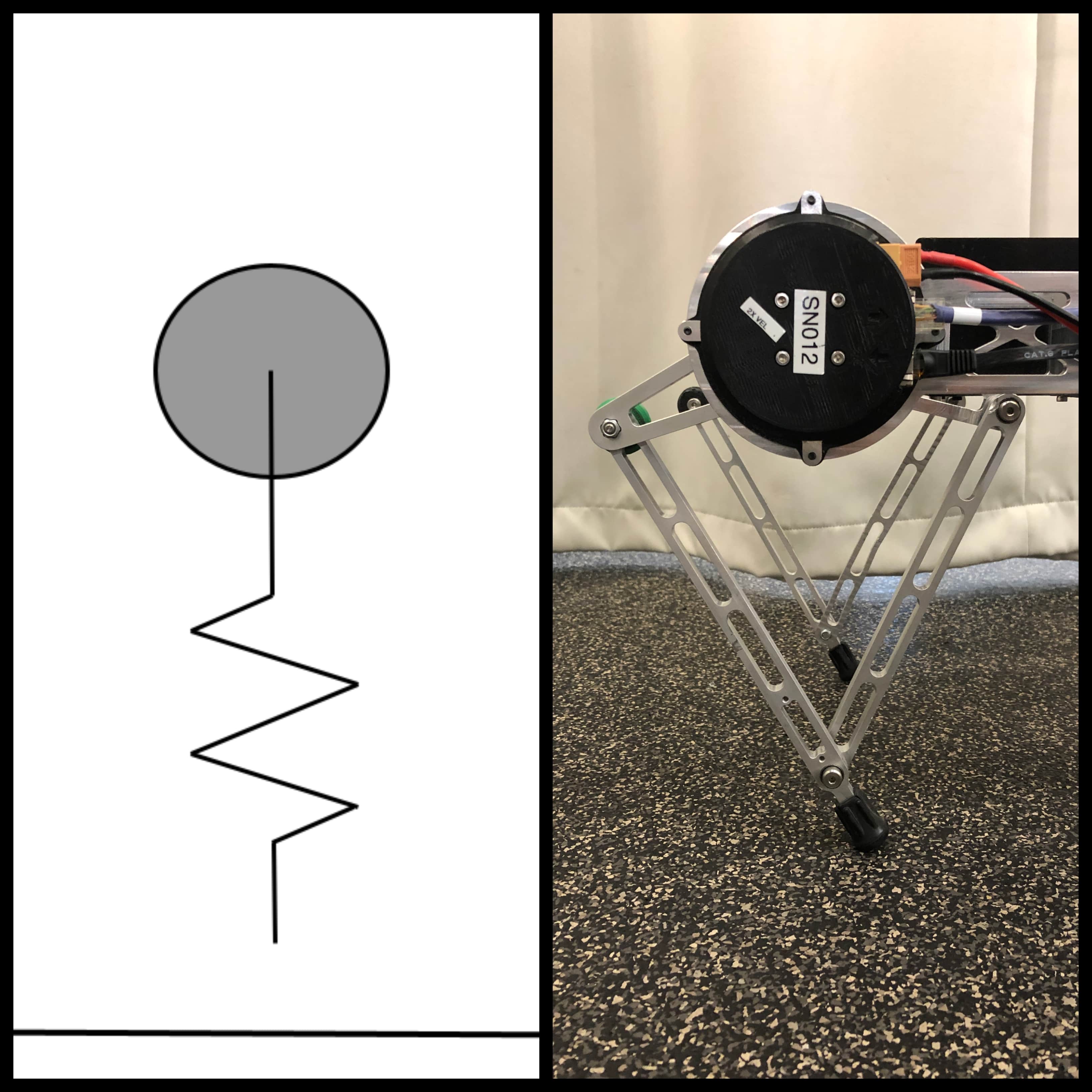}
    \caption{Left: A 1DoF Vertical Hopper, which consists for a round mass and a spring leg; Right: A 2DoF Minitaur limb, which consist of two actuators at the hip and an aluminum leg}
    \label{fig:v_hop}
\end{figure}
Section \ref{sec:model_anaylsis} presents a motivating preview of our hybrid fore-aft pronking controller by recourse to a previously developed \cite{de_parallel_2015} monopedal hopping template controller \cite{full_templates_1999}. Section \ref{sec:tripronk} describes the empirical anchoring \cite{full_templates_1999} of this monopedal template in a three-legged (single limb-lost) quadrupedal robot and documents its empirically stable, repeatable tripedal pronking. Section \ref{sec:conclusion} summarizes the findings and provides a brief discussion on future research directions.
\section{Model Analysis: SLIP}\label{sec:model_anaylsis}
\subsection{1DoF Vertical Hopping}
To achieve continuous vertical hopping, a controller is needed that exerts enough force to keep the system hopping, despite energy loss due to resistive forces. The vertical hopping behavior is made up of a (ballistic) flight phase and a stance phase. In flight, gravity exerts a constant downward acceleration, as modeled by the familiar system:

\begin{equation}
    \label{eqn:flight}
    \ddot{\chi} + \gamma = 0
\end{equation}

where $\chi$ is the spring deflection and $\gamma$ is the mass-specific gravitational acceleration. In stance, in the absence of the active damping (AD) controller to be reviewed below, the mass-spring-damper model (also referred to as the vertical hopping template \cite{full_templates_1999} dynamics is
% removed rho
\begin{equation}
    \label{eqn:stance}
    \ddot{\chi}+ \frac{c}{\mu}  \Dot{\chi} + \frac{k}{\mu} \chi = \tau
\end{equation}
where $\mu$ is the mass, $c$ is the damping constant, $k$ is the spring constant, and $\tau$ is the (mass-normalized) input torque.

For this application, we have adapted a controller in which the actuator imposes forces that emulate a damped spring, whose sign varies over the stance phase, so as to introduce or dissipate energy as required to regulate hopping height. This concept was first introduced for monopedal running \cite{secer_control_2013}, then extended to bipedal fore-aft hopping with tail actuation \cite{de_parallel_2015}, and further used in the control of ``virtual bipedal gaits" for a quadruped robot
\cite{de_vertical_2018}. Following \cite{de_parallel_2015}, a change of coordinates is introduced, where $x_1 := \chi$, $x_2 := \frac{\dot\chi}{\omega}$, and $\mathbf{x} := [x_1, x_2]^{\intercal}$, resulting in:
\begin{equation}
    \label{eqn:first_order_dyn}
    \mathbf{\Dot{x}} = -\omega \begin{bmatrix}
    0 & -1 \\
    1 & 0 
    \end{bmatrix} \mathbf{x} + \begin{bmatrix}
    0 & 1  
    \end{bmatrix}(-2 \bar\beta \omega x_2 + \frac{\tau}{\omega})
\end{equation}
Here, the physical liftoff and touchdown events ($\chi = 0$) correlate to the hybrid reset events at $x_1=0$. 

The active damping control law,
\begin{equation}
    \label{eqn:controller}
    \tau := k_{t} \cos{\angle \mathbf{x}}
\end{equation}
results in the closed-loop stance dynamics,
\begin{equation}
    \label{eqn:closed_loop}
    \mathbf{\Dot{x}} = -\omega \text{J} \mathbf{x} + e^{\intercal}_{2}(-2 \bar\beta \omega x_2 + \frac{k_{t} \cos{\angle \mathbf{x}}}{\omega})
\end{equation}
where $\angle \mathbf{x} = \arctan \left(\frac{x_1}{x_2}\right)$, $\Bar{\beta} = \frac{c}{2 \mu \omega}$, and $\omega = \sqrt{\frac{k}{\mu}}$.
\subsection{Controlled Fore-aft Speed}
The comprehensive, longstanding literature on non-disabled running animals \cite{blickhan_full_1993} establishes that their ground reaction forces closely resemble those of a bouncing pogo stick \cite{dickinson_how_2000} or an inverted pendulum \cite{cavagna_mechanical_1977,saranli_toward_1998}. In this work, we build off \cite{ghigliazza_simply_2003}, which showed that a fixed touchdown angle can admit a reasonable basin of stability around an emergent attracting steady-state velocity in SLIP.
\section{Dynamical Movement: Tripedalism}\label{sec:tripronk}
In this section, the anchoring of the previously mentioned 2DoF SLIP template by a spatial quadrupedal robot will be presented and its empirical performance documented.
\subsection{From four legs to three legs}
For a quadruped with functional limbs, the coordination of two independent, mechanically coupled 1DoF vertical hopping models comprises the ``slot-hopper'' template that encodes stable representatives of all virtual biped gaits  including pronking \cite{de_vertical_2018}, the focus of this work. Under the assumption of symmetry in the transverse and sagittal planes, the stability of limit cycles obtained from the formal analysis of the slot hopper has been shown to coincide with the execution of steady state gaits by physical quadrupeds \cite{de_vertical_2018, Greco_Koditschek_2023}.

For a quadruped with an amputated limb, this template cannot provide the same stability guarantees as previously discussed. Most critically, the assumption of body symmetry is invalidated, since the morphology of the robot now consists of an odd-number of limbs. Moreover, in this work, we strive to maintain a statically stable pose at touchdown and liftoff, requiring that each of the remaining limbs is assigned a rest length distinct from the others to compensate for the lost limb. Finally, during stance phase, each of the remaining limbs is assigned an angle relative to the body distinct from the others so as to eliminate the load on the one missing. Having abandoned the slot hopper template \cite{de_vertical_2018} for these reasons, we instead follow in the tradition of Raibert \cite{raibert_legged_1986}, approaching the coordination of the three functional limbs by treating them as a single virtual limb. Specifically, we hypothesize that the appropriately coordinated control of three legs on a physical robot can anchor \cite{full_templates_1999} the behavior of a planar body forced by a single ``virtual'' leg using the AD controller introduced in Sec. \ref{sec:model_anaylsis}.

\subsection{Experimental Setup}
The experimental platform for this work is the Ghost Robotics Minitaur (Fig. \ref{fig:pronk} \cite{ghost_robotics}), an underactuated, direct-drive quadrupedal robot. Each of its limbs consists of a 5-bar linkage with two actuated DoF capable of independently controlling toe extension and hip flexion, with no ab/adduction joints and no passively compliant elements \cite{kenneally_design_2016}. All experiments were conducted indoors on linoleum flooring. Empirical measurements were obtained simultaneously from microcontroller logging and motion capture (mocap) tracking\footnote{Raw data was filtered in MATLAB using the ``moving median'' smoothing method with window length set to 8 \cite{mathlab_smooth_data}.}.

\subsection{Experimental Results}
\subsubsection{Quasi-static stability via pose adjustment}
To test its ability to perform dynamic tripedal locomotion, Minitaur was tasked with closed-loop vertical pronking. To mimic an amputated limb, the front-right limb\footnote{Any leg could have been assigned the missing role. Considering that the weight distribution of the non-disabled Minitaur is greater toward the rear, the removal of the front limbs was found to be easier and safer. The robot's direct drive actuators \cite{kenneally_design_2016}, which generally tend to overheat during high energy gaits. We found retracting the front limbs allowed the robot to complete multiple trials of vertical pronking without quickly burning the motors.} is rendered \textit{missing} --- set to a fixed, retracted position such that it does not make contact with the ground. At rest, removing one limb's contact with the ground alters the robot's polygon of support and places its center of mass (CoM) outside of that area, prompting the body of the robot to fall onto the retracted limb.  

To shift the robot's CoM into its new polygon of support (in other words, to become quasi-statically stable at rest), the remaining functional legs must redistribute the weight of the robot away from the missing limb. Absent ad/abduction actuation, limb kinematics restrict the remaining toes to move in the robot's sagittal plane, motivating the following re-balancing strategy. At rest, the limb ipsilateral and the limb contralateral to the missing limb are set to the same desired length, while the limb diagonal to the missing limb is significantly decreased hereby redistributing the robot's weight to bring the projected mass center into the interior of the toes’ support polygon. Similarly, during dynamical movements, reducing the length and force of the limb diagonally opposed to its missing counterpart relative to the other two remaining functional limbs guards against body tip over  during liftoff and touchdown events.
\subsubsection{Fore-aft Pronking with Three Legs} 
To demonstrate tripedal pronking, Minitaur starts on four functional limbs in a quasi-statically stable pose. Upon command, the robot retracts the designated missing limb, shifts to the statically stable three-legged pose and initiates the virtual monopedal active damping control strategy we now detail.

For the physical implementation, each functional limb is treated as a single actively damped mass-spring-damper system. For limb j\footnote{Note the limbs are labeled left-front,left-back (LB), right-front (RF), and right-back (RB). The angle displacement is the distance from the negative y Euler angle as defined in the Ghost Robotics SDK \cite{ghost_robotics}}, the commanded feedback policy is:

% removed extra js'
\begin{equation}
    \label{eqn:closed_loop_robot}
    u_j := -2 \Bar{\beta}_{s} \omega_{ss} \Dot{\chi}_j - \omega_{ss}^{2} (\chi_j-\rho_j) + k_{st}  \cos{\angle (\bar\chi, \bar{\dot{\chi}})}
\end{equation}

where $\angle (\bar\chi,\bar{\dot{\chi}}) = \arctan \left(\frac{\bar{\chi}}{\bar{\dot\chi}}\right)$, $\bar{\chi}$ and $\bar{\dot{\chi}}$ denote the averaged extension positions and velocities of the three remaining legs, and $\omega_{ss} = \sqrt{\frac{k_{ss}}{\mu_r}}$ (see Table \ref{table_ctrl} for a  list of the parameter values used in experiments). Note that whereas the average limb position and average limb velocity are used in the active damping term, the individual limb positions and velocities are used in the spring and damping terms of this feedback law. We discovered empirically that this combination of local and distributed feedback disrupted the preflexively stable bounding behavior described in \cite{de_vertical_2018}, resulting in   the desired stable pronking gait for reasons that are presently under investigation.

Inspired by the fixed angle stepping policy \cite{ghigliazza_simply_2003}, the desired fore-aft velocity is adjusted by setting each leg angle to empirically chosen  fixed values at touchdown (td) and liftoff (lo) as listed in Table \ref{table_ctrl}\footnote{For the quadrupedal fore-aft pronking gait, the leg angle at touchdown and liftoff is the same.}. The leg angles are reset using the Ghost Robotics SDK proportional-derivative (PD) controller, using a P gain of $2.0$ and a D gain of $0.03$. When the robot is lifting off, $\theta_{j_{lo}}$ is set such that the body goes into aerial phase in the desired fore-aft direction. When the robot touches down, $\theta_{j_{td}}$ is set such that the limbs are directly under the body to prevent it from making contact with the ground. To adjust the fore-aft speed, the AD gain must be altered, which affects the amount of force exerted by each limb against the ground. A large applied force in the fore-aft direction results in covering a distance in a shorter amount of time than a lower force.

\begin{figure}[htbp]
    \centering
    \begin{subfigure}{0.46\textwidth}
    \includegraphics[width=\textwidth]{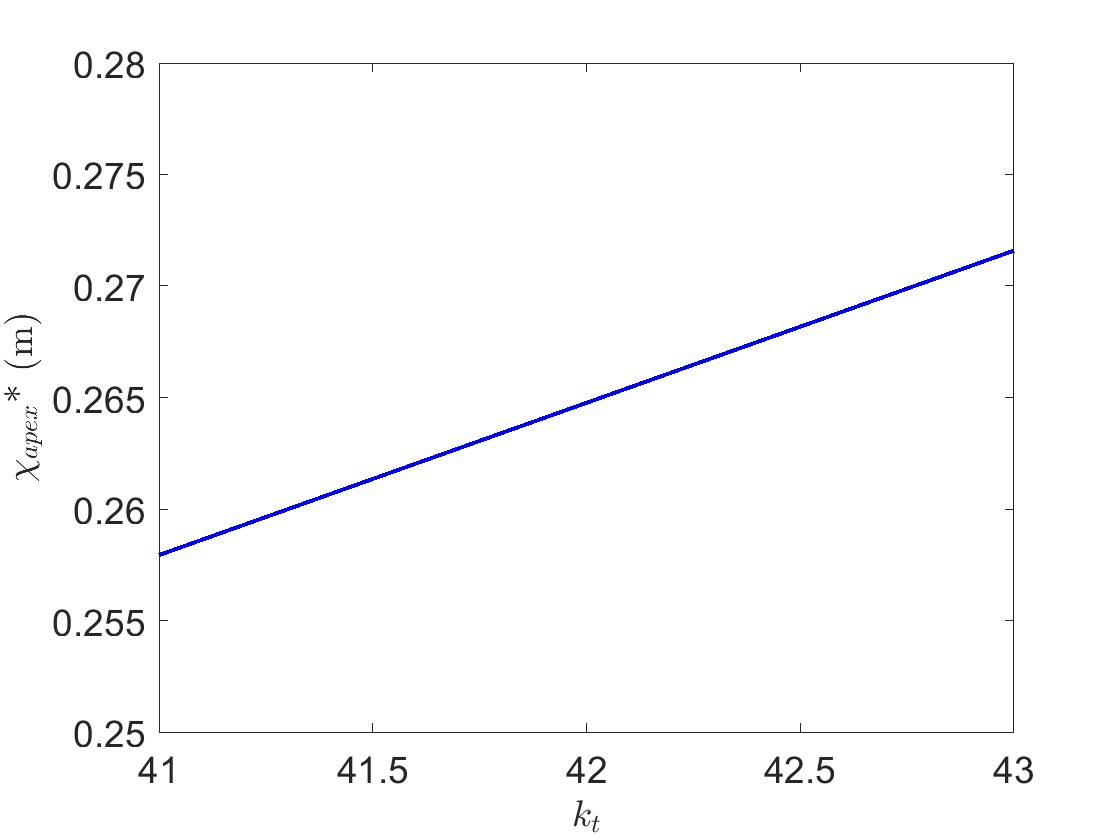} 
    \caption{}
    \label{fig:kt1}
    \end{subfigure}
    \begin{subfigure}{0.46\textwidth}
    \includegraphics[width=\textwidth]{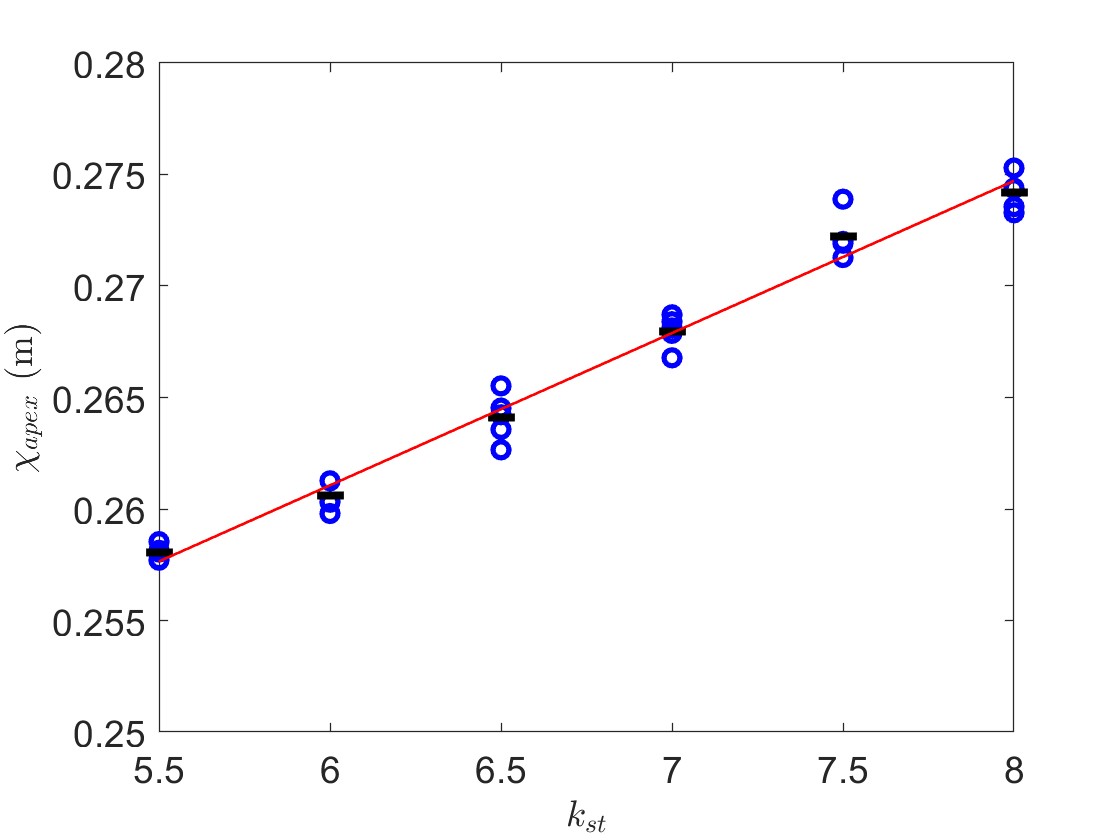}
    \caption{}
    \label{fig:kt2}
    \end{subfigure}
    \caption{Steady state hopping height as a function of the active damping gain: (a) Graph of the fixed point (in apex coordinates) of the hybrid dynamical closed loop (\ref{eqn:closed_loop}) as a function of $k_t$ (derived in \cite{de_parallel_2015}); (b) Steady state hopping height (estimated from mocap tracking data) of the hopping robot,  plotted for five different trials (open blue circles) at seven different set points of the gain, ${k}_{st}$. The means at each set point (black bars) exhibit a roughly affine relationship as predicted by the active damping vertical hopping template. The line of best fit is in red.}
    \label{fig:img6}
\end{figure}

Mathematical analysis detailed in \cite{de_parallel_2015} demonstrates that the steady state hopping height of the 1DoF closed loop  hybrid dynamical system resulting from (\ref{eqn:controller}) is an affine function (plotted in Fig. \ref{fig:kt1}) of the vertical gain, $k_t$. To test our hypothesis that the actively damped monoped template (\ref{eqn:closed_loop}) is anchored by the controller described above (\ref{eqn:closed_loop_robot}), we ran a series of hopping trials varying the software (vertical) gain, ${k}_{st}$, applied to the physical robot. Vertical tripedal pronking was tested for several distinct values of ${k}_{st}$ with the results plotted in Fig. \ref{fig:kt2}. Each experiment consists of five trials, during which the unconstrained robot performed $20$ consecutive vertical pronks with $5$ minutes rest in between each trial to allow the motors to cool.

Fore-aft pronking was tested for several distinct values of ${k}_{st}$. Each experiment consisted of three trials, during which the unconstrained robot pronked a distance of $2$ meters with $10$ minutes rest in between each trial to allow the motors to cool. For a comparison analysis, we also conducted the same fore-aft pronking experiments on the robot with four functional limbs. For each trial, an estimate of the cost of transport \cite{Gregorio_Ahmadi_Buehler_1997} was computed as,
\begin{equation}\label{eqn:specific}
    \sigma : = \frac{\bar{P}}{\mu_{r} g \bar{v}}
\end{equation}
where $\bar{P}$, an estimate of the average total power expended, is calculated using a procedure detailed in the appendix, and $\bar{v}$, the average speed, is the ratio of distance traveled to elapsed trial time. 
\begin{table}[ht]
\caption{Model Parameters}
\label{table_sim}
\begin{center}
\begin{tabular}{c c c}
\hline
Symbol & Parameter (Units) & Value\\
\hline
\rule{0pt}{3ex}
$\chi_{0}$ (Eqn. \ref{eqn:stance}) & Initial Position of Mass ($m$) & $0.28$\\

$\rho$ (Eqn. \ref{eqn:stance}) & Spring Rest Length ($m$) & $0.18$\\

$\mu$ (Eqn. \ref{eqn:closed_loop}) & Mass ($kg$) &  $6.173$\\

$k$ (Tbl. \ref{eqn:stance}) & Spring Const. ($N/m$) & $1500$\\

$\gamma$ (Eqn. \ref{eqn:flight}) & Gravitational Force ($m/s^2$) &  $9.81$\\

$c$ (Tbl. \ref{eqn:stance}) & Damping Const. & $3.2$\footnote{for Figure \ref{fig:img6} the damping const. was increased to $11.9$}\\
\hline
\end{tabular}
\end{center}
\end{table}
%
%
%
% The vertical hopping template (eqn here) has a unique attracting periodic orbit.
% increased the font size
\subsubsection{Data Explained}
\begin{figure*}[htbp]
    \centering
    \begin{subfigure}{0.37\textwidth}
    \centering
    \includegraphics[width=\textwidth]{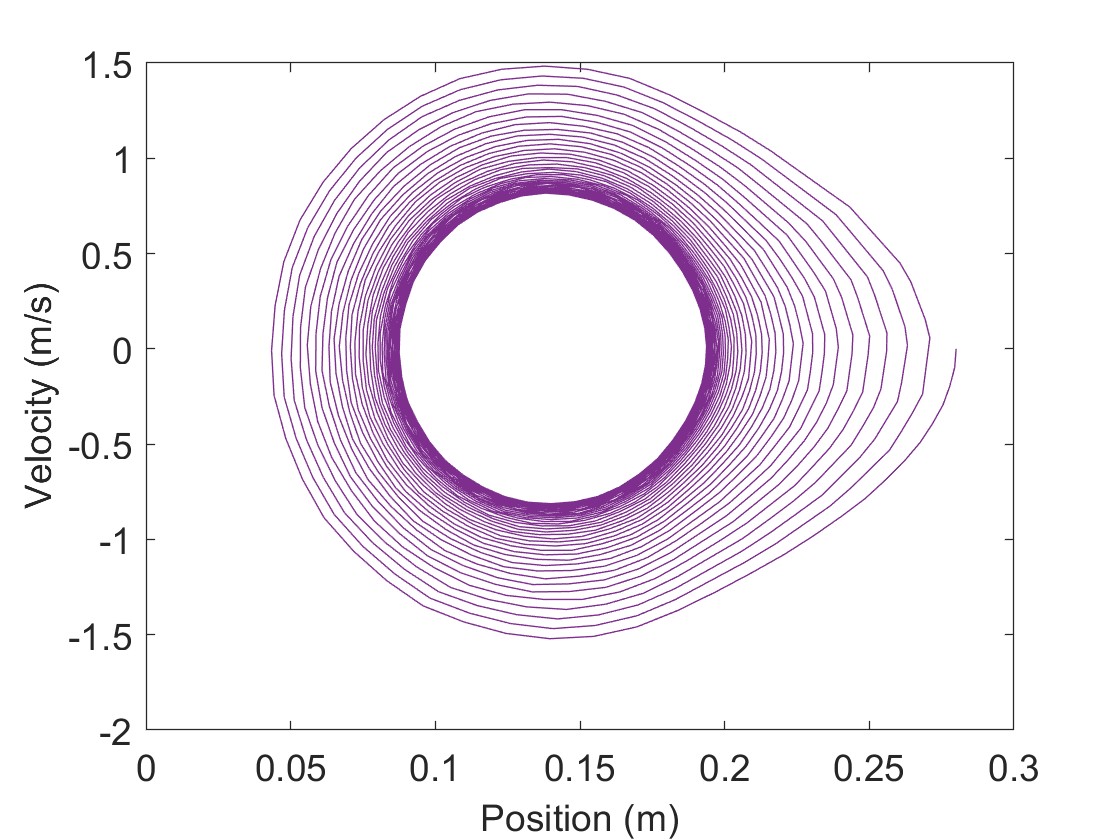} 
    \caption{Numerical Simulation
    \\$k_t=5.5$}
    \label{fig:subim1}
    \end{subfigure}
    \begin{subfigure}{0.37\textwidth}
    \centering
    \includegraphics[width=\textwidth]{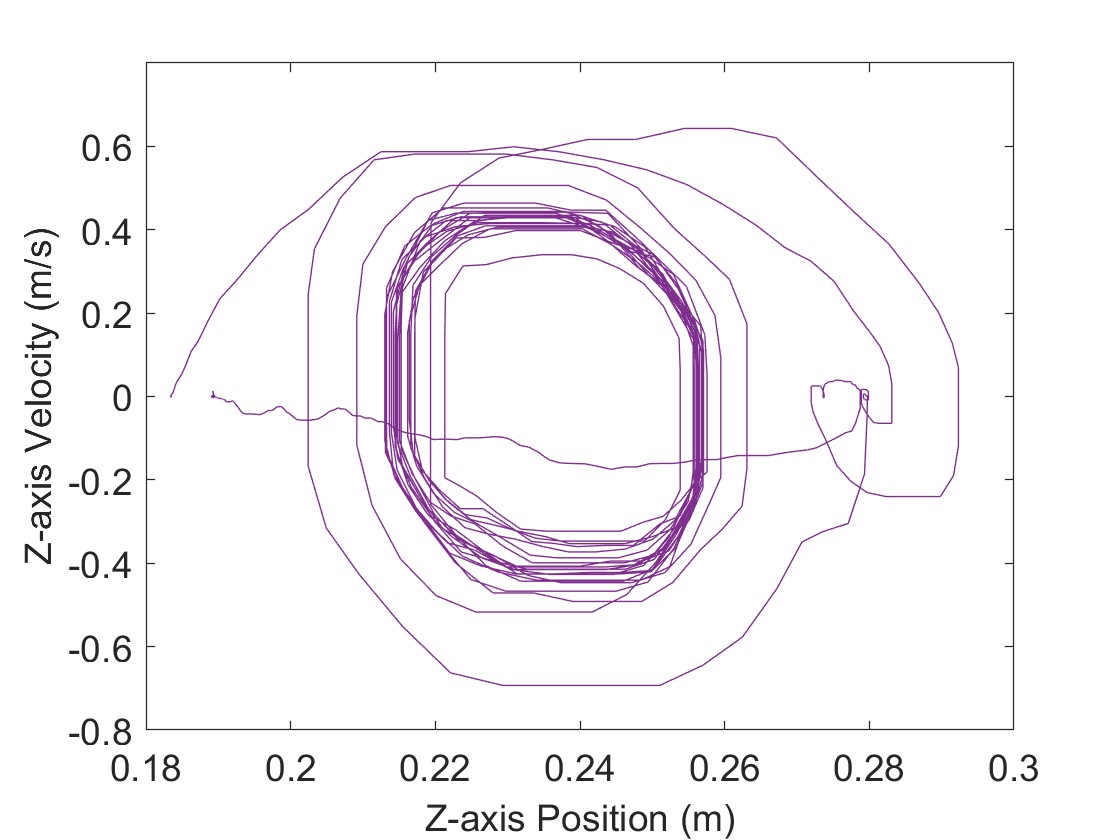}
    \caption{MoCap Tracking
    \\$k_{st}=5.5$}
    \label{fig:subim2}
    \end{subfigure}
    \centering
    \begin{subfigure}{0.37\textwidth}
    \centering
    \includegraphics[width=\textwidth]{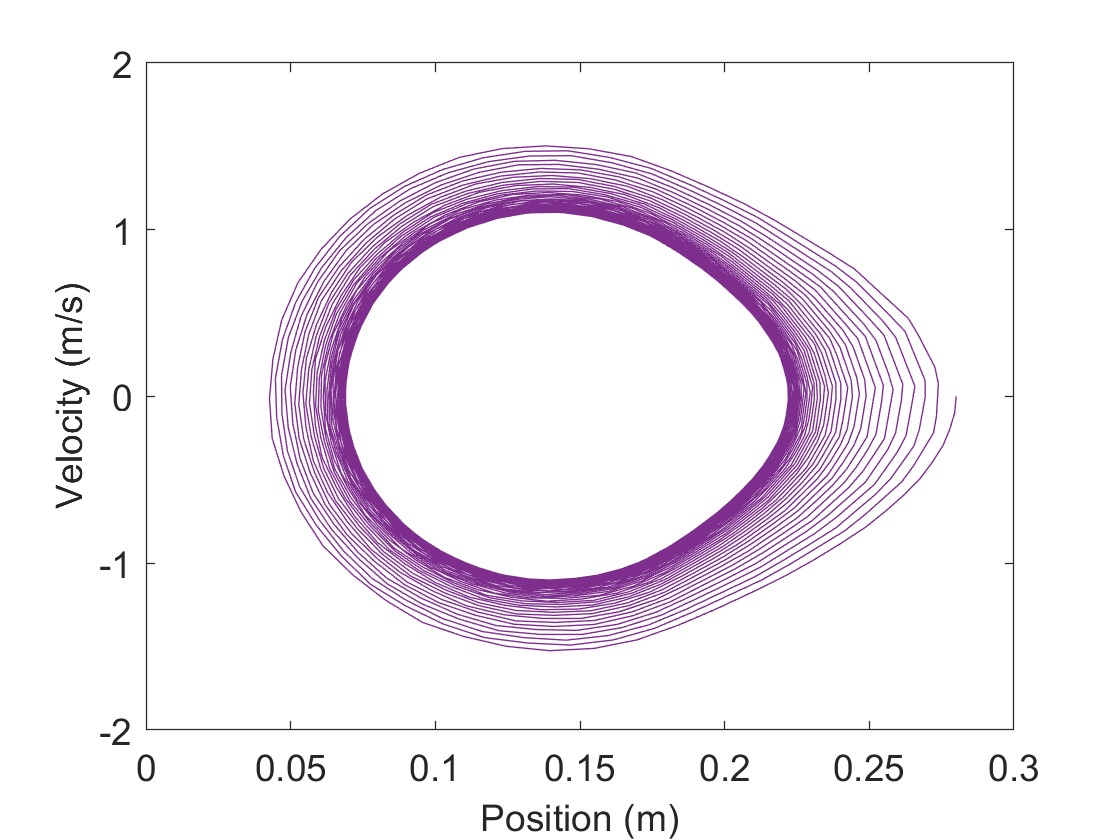} 
    \caption{Numerical Simulation \\$k_t=7.5$}
    \label{fig:subim3}
    \end{subfigure}
    \begin{subfigure}{0.37\textwidth}
    \centering
    \includegraphics[width=\textwidth]{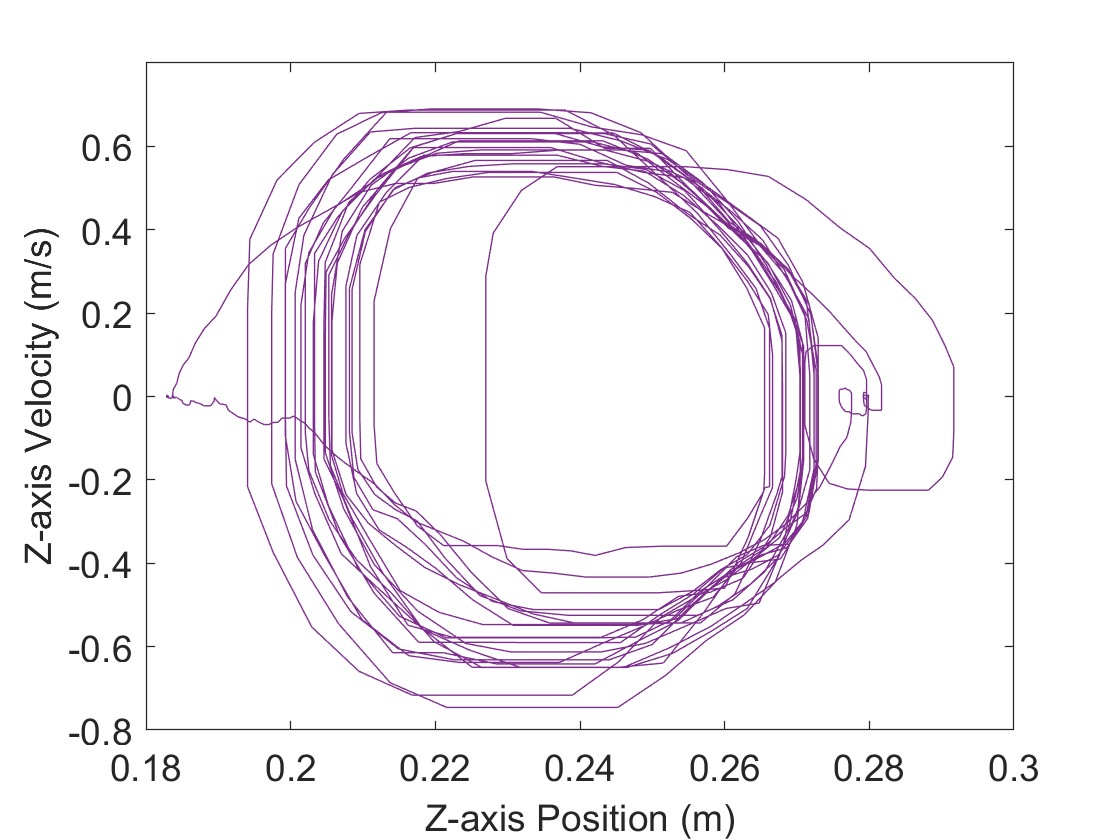}
    \caption{MoCap Tracking
    \\$k_{st}=7.5$}
    \label{fig:subim4}
    \end{subfigure}
    \caption{Limit cycles for different vertical gains: (a;c) the actively damped vertical hopper simulated in MATLAB; (b;d) Minitaur performing vertical tripedal pronking --- data retrieved from motion capture tracking.}
    \label{fig:img2}
\end{figure*}

\begin{figure}[htbp]
    \centering
    \begin{subfigure}{0.46\textwidth}
    \includegraphics[width=\textwidth]{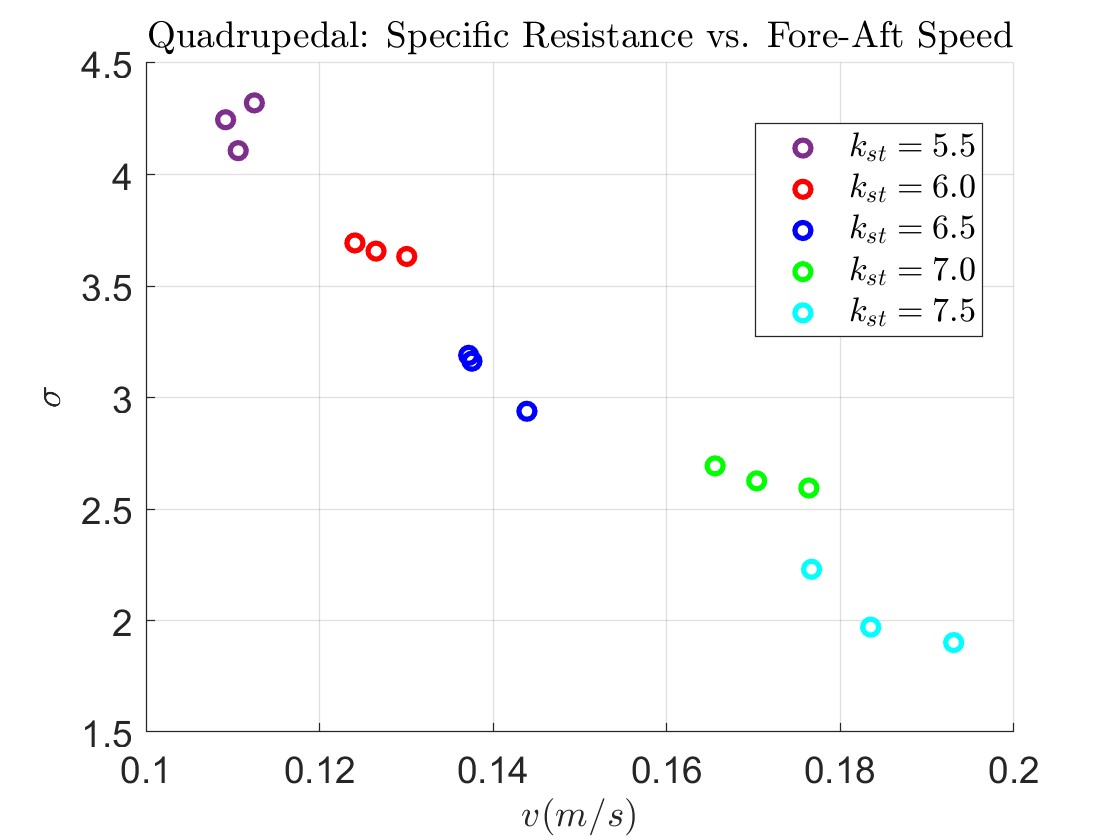} 
    \caption{}
    \label{fig:sig1}
    \end{subfigure}
    \begin{subfigure}{0.46\textwidth}
    \includegraphics[width=\textwidth]{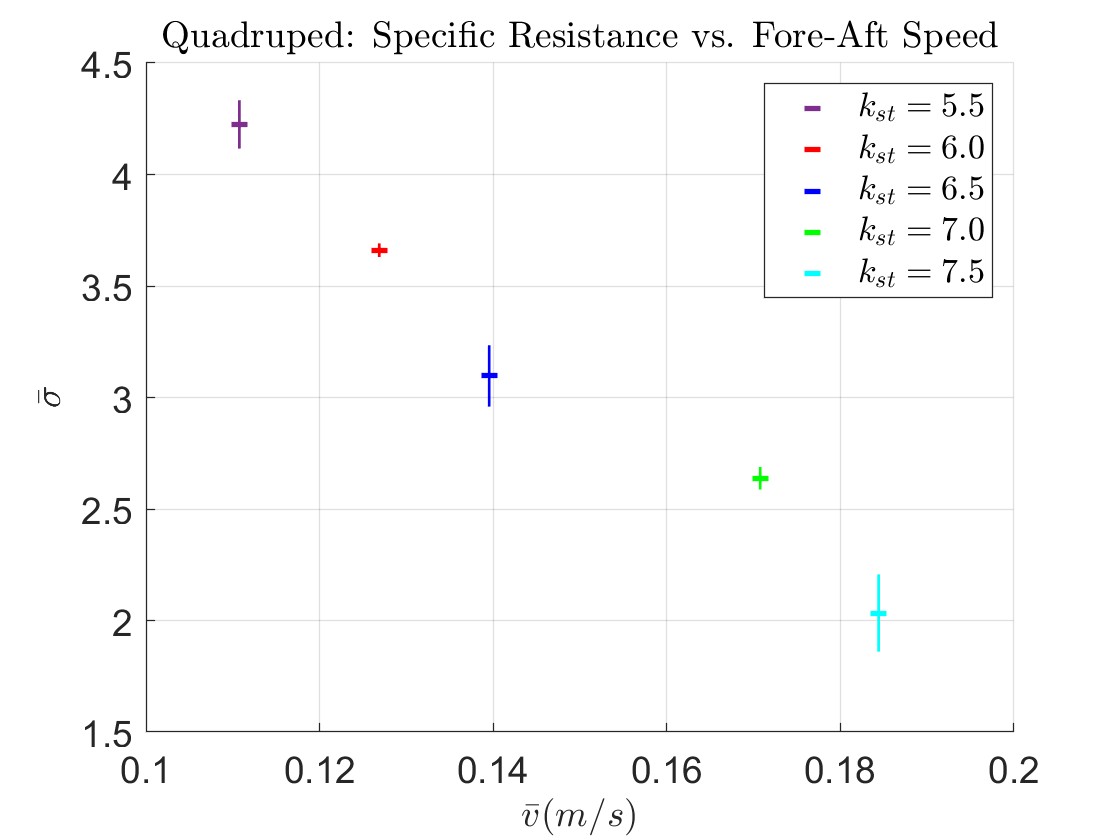}
    \caption{}
    \label{fig:sig2}
    \end{subfigure}
    \caption{Quadrupedal Fore-aft Pronking: (a) Specific resistance (\ref{eqn:specific}) computed for each trial over the fore-aft velocity of each trial, as a function of the active damping gain; (b) Average fore-aft velocity plotted against average specific resistance over all three trials as a function of the active damping gain. The horizontal lines represent the mean and the vertical lines are the standard deviation.}
    \label{fig:img}
\end{figure}

\begin{figure}[htbp]
    \centering
    \begin{subfigure}{0.46\textwidth}
    \includegraphics[width=\textwidth]{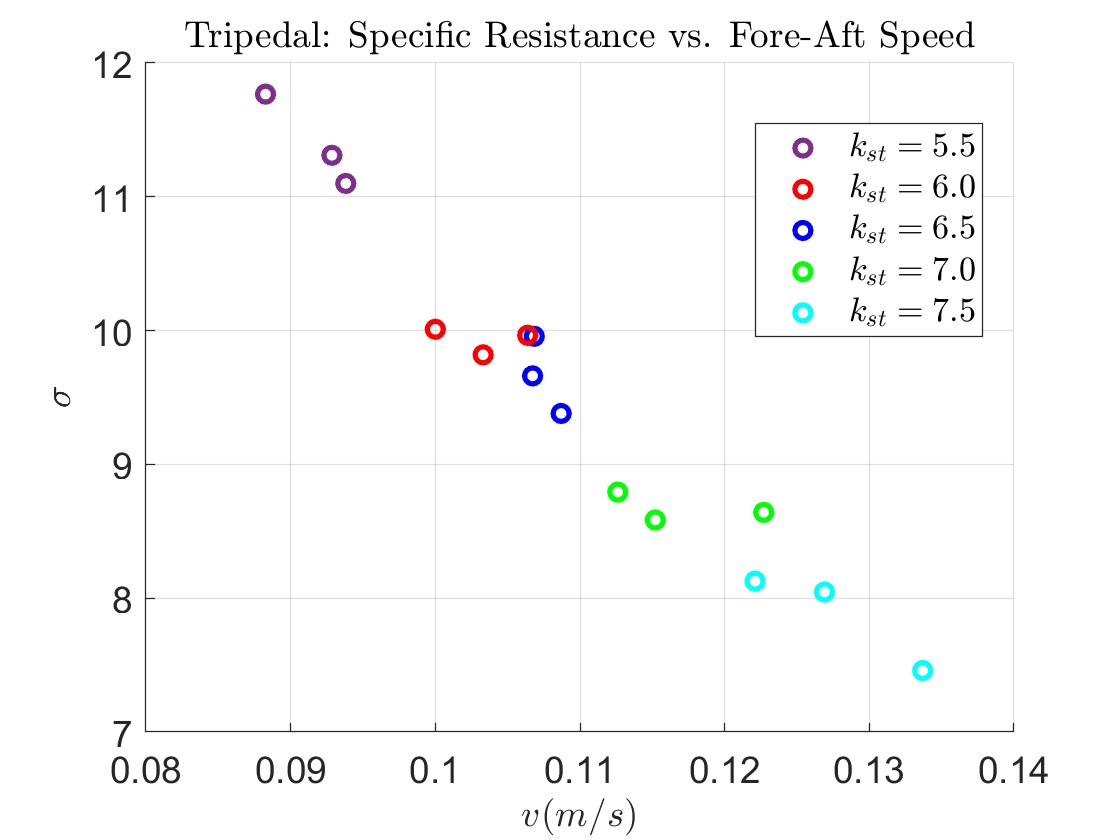} 
    \caption{}
    \label{fig:sigm1}
    \end{subfigure}
    \begin{subfigure}{0.46\textwidth}
    \includegraphics[width=\textwidth]{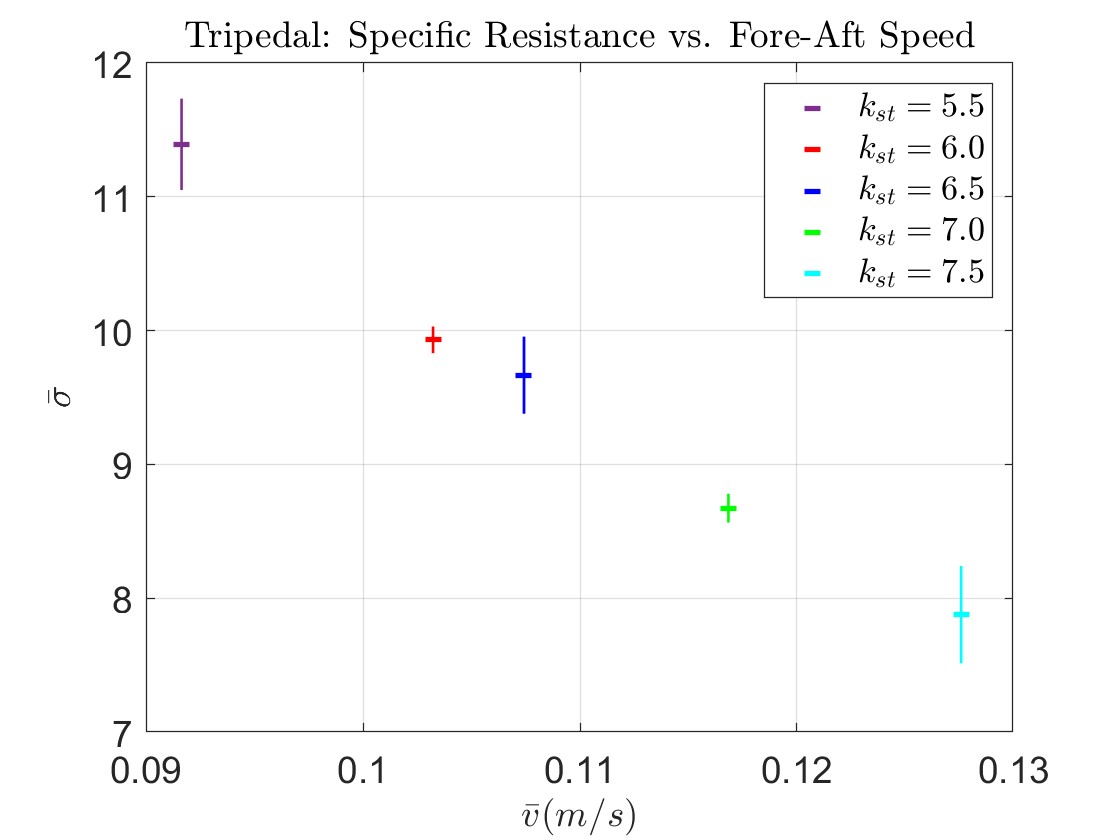}
    \caption{}
    \label{fig:sigm2}
    \end{subfigure}
    \caption{Tripedal Fore-aft Pronking: (a) Specific resistance (\ref{eqn:specific}) computed for each trial over the fore-aft velocity of each trial, as a function of the active damping gain; (b) Average fore-aft velocity plotted against average specific resistance over all three trials as a function of the active damping gain. The horizontal lines represent the mean and the vertical lines are the standard deviation.}
    \label{fig:imgg}
\end{figure}

To corroborate the applicability of the stable active damping controller to this new physical setting, a comparison of the numerically simulated results and the empirical data is presented. As validation that steady state has been reached, we examined the limit cycles for the numerical and empirical data. The model parameters for the simulation are given in Table \ref{table_sim}. The dimensions of the 2DoF SLIP model directly corresponds to the actuator-leg arrangement for a single limb on Minitaur. The numerical simulation begins in stance phase, with the vertical hopper initially positioned at rest length. The control parameters for the physical implementation of the tripedal and quadrupedal gaits are given in Table \ref{table_ctrl} and Table \ref{table_ctrl_quad} respectively. Figure \ref{fig:img6} is a plot of the average hopping height of the robot during steady state tripedal pronking\footnote{Because the individual leg controllers (\ref{eqn:closed_loop_robot}) entail a mixture of individual and averaged leg data, we have not yet determined the best way to calibrate the software gain parameters. Hence we have split out the graph of the model's apex height formula (as a function of the modeled $k_t$) in Fig. \ref{fig:kt1} from the plot of the measured apex data (as a function of the commanded $k_{st})$ in Fig. \ref{fig:kt2}.} As the commanded vertical gain, $k_{st}$, is
increased, the average apex position monotonically increases with an affine best fit (the red line in Fig. \ref{fig:kt2}) whose slope reasonably well fits that of the model (Fig. \ref{fig:kt1}). Figures \ref{fig:subim2} and \ref{fig:subim4} display the limit cycles for a single trial of vertical hopping with a ``vertical" gains of $5.5$ and $7.5$, respectively\footnote{The height of the robot while at rest with limb length $0.18$m is $0.247$m}. As the vertical gain, $k_t$, is increased, the average apex position monotonically increases in a roughly affine manner, which agrees with the formal model previously described. Fig. Figure \ref{fig:img} shows the individual and aggregate CoT as a function of fore-aft velocity for walking and pronking with fixed angle stepping for the intact robot. Figure \ref{fig:imgg} presents the individual and aggregate CoT as a function of fore-aft velocity for walking and pronking with fixed angle stepping for the three-legged (damaged) robot. These data corroborate the expected pattern of decreasing cost of transport with modestly increasing speeds for both the intact quadruped and its three-legged counterpart while confirming the intuition that the loss of a limb incurs substantially greater (by roughly by a factor of two) cost at lower (by roughly a factor of 3/4) speed. 
\begin{table}[ht]
\caption{Control Parameters for Tripedal Gaits}
\label{table_ctrl}
\begin{center}
\begin{tabular}{c c c c}
\hline
Gait & Symbol & Parameter (Units) & Value\\
\hline
\rule{0pt}{3ex}
Both & $\bar\beta_s$ (Eqn. \ref{eqn:closed_loop_robot}) & Software Damping & $0.25$ \\
& $\mu_r$ (Eqn. \ref{eqn:closed_loop_robot}) & Mass of Robot ($kg$) &  $6.173$ \\
\hline
\rule{0pt}{3ex}
Vertical & $k_{ss}$ (Eqn. \ref{eqn:closed_loop_robot}) & Software Spring Const. & $1500$\\
& $\rho_{LF}$ (Eqn. \ref{eqn:closed_loop_robot}) &  LF Rest Length ($m$) & $0.17$ \\
& $\rho_{LB}$ (Eqn. \ref{eqn:closed_loop_robot}) &  LB Rest Length ($m$) & $0.138$ \\
& $\rho_{RB}$ (Eqn. \ref{eqn:closed_loop_robot}) &  RB Rest Length ($m$) & $0.18$ \\
& $\theta_{LF}$ &  LF Angle  ($rad$) & $0.0$ \\
& $\theta_{LB}$ &  LB Angle  ($rad$) & $-0.15$ \\
& $\theta_{RB}$ &  RB Angle  ($rad$) & $+0.2$ \\
\hline
\rule{0pt}{3ex}
Fore-aft & $k_{ss}$ (Eqn. \ref{eqn:closed_loop_robot}) & Software Spring Const. & $1300$\\
& $\rho_{LF/RB}$ (Eqn. \ref{eqn:closed_loop_robot}) &  LF,RB Rest Length ($m$) & $0.18$ \\
& $\rho_{LB}$ (Eqn. \ref{eqn:closed_loop_robot}) &  LB Rest Length ($m$) & $0.145$ \\
& $\theta_{{LF}_{lo}}$ &  LF Angle ($rad$) & $-0.15$ \\
& $\theta_{{LB}_{lo}}$ &  LB Angle ($rad$) & $-0.1$ \\
& $\theta_{{RB}_{lo}}$ &  RB Angle ($rad$) & $+0.15$ \\
& $\theta_{{LF}_{td}}$ &  LF Angle ($rad$) & $-0.1$ \\
& $\theta_{{LB}_{td}}$ &  LB Angle ($rad$) & $-0.05$ \\
& $\theta_{{RB}_{td}}$ &  RB Angle ($rad$) & $+0.1$ \\
\hline
\end{tabular}
\end{center}
\end{table}
%
% updated graphs with new legends and colors; increased font size; add mass to parameter table
\begin{table}[ht]
\caption{Control Parameters for Quadrupedal Gait}
\label{table_ctrl_quad}
\begin{center}
\begin{tabular}{c c c c}
\hline
Gait & Symbol & Parameter (Units) & Value\\
\hline
\rule{0pt}{3ex}
Fore-aft & $\Bar{\beta}_s$ (Eqn. \ref{eqn:closed_loop_robot}) & Software Damping & $0.25$ \\
& $\mu_r$ (Eqn. \ref{eqn:closed_loop_robot}) & Mass of Robot ($kg$) &  $6.173$ \\
& ${k}_{ss}$ (Eqn. \ref{eqn:closed_loop_robot}) & Software Spring Const. & $800$\\
& $\rho_{LF,LB,RF,RB}$ (Eqn. \ref{eqn:closed_loop_robot}) & Rest Length ($m$) & $0.18$ \\
& $\theta_{LF,LB,RF,RB}$ & Angle ($rad$) & $-0.05$ \\
\hline
\end{tabular}
\end{center}
\end{table}
\section{CONCLUSION}\label{sec:conclusion}
This paper presents what we believe is the first demonstration of a stable dynamical gait for a quadrupedal robot missing one of its limbs. Specifically, our empirical results suggest that the active damping \cite{de_parallel_2015} monopedal vertical hopping template (\ref{eqn:closed_loop}) with fore-aft speed selected by a variant of fixed angle stepping \cite{ghigliazza_simply_2003} can be anchored in the quadruped's remaining three limbs coordinated to perform a pronking gait. Moreover, while our preliminary specific resistance measurements confirm the intuition that this already inefficient gait is considerably more costly to implement in the face of a lost limb, the rough doubling of specific resistance and reduced speed nevertheless seems far preferable to the body-slamming consequences almost surely incurred by any quasi-static locomotion alternative. 

This initial exploration of stable steady state fore-aft tripedal pronking by a damaged quadrupedal robotic platform is the stepping stone toward work in progress targeting more efficient stable fault tolerant dynamic gaits for quadrupeds that encounter permanent limb damage while operating in unstructured outdoor environments. At the same time, more sophisticated stepping controllers should yield control of fore-aft speed independent of the hopping height. In parallel, future work will investigate online learning methodologies for limb fault detection in legged  robots. 
%\addtolength{\textheight}{-12cm}   % This command serves to balance the column lengths
                                  % on the last page of the document manually. It shortens
                                  % the textheight of the last page by a suitable amount.
                                  % This command does not take effect until the next page
                                  % so it should come on the page before the last. Make
                                  % sure that you do not shorten the textheight too much.

%%%%%%%%%%%%%%%%%%%%%%%%%%%%%%%%%%%%%%%%%%%%%%%%%%%%%%%%%%%%%%%%%%%%%%%%%%%%%%%%
\section*{APPENDIX} \label{appendix}
\subsection{Implementing an Active Damping Controller using the Ghost SDK}\label{appendix:a}

Ghost Robotics provides a software development kit (SDK) that allows you to control and develop behaviors for their robots and motor modules. The Ghost Robotics SDK \cite{ghost_robotics} provides a class to coordinate multiple joints together into a manipulator (e.g. a leg) with specified kinematics. For Minitaur, each limb manifests itself as a 5-bar symmetric linkage consisting of two actuated joints. Within the Limb class, there are two parameters, angle and extension, (referred to as end-effector coordinates in the SDK documentation) that commands the movements of each limb. The extension coordinate is used to extend the limb away from the body or compress the limb toward the body. The angle coordinate governs the rotation of the limb toward the front or rear of the body. 

To illustrate vertical tripedal pronking, the control strategy, outlined in (\ref{eqn:closed_loop_robot}), is the input for the Open Loop function of the Limb class. This function uses voltage control on the selected end-effector coordinate (e.g. extension) of the limb. The force outputted from this function commands each active limb to extend (essentially pushing against the ground to lift the body upward). If a large enough force is exerted, the robot's toes will break contact with the ground. To restrict the robot to pronking in the vertical plane, we used the SDK's proportional-derivative (PD) controller --- with gains of 2.0 and 0.03, respectively --- to set each limb angle to a fixed position.

\subsection{Total Power Consumption}
Ideally, we would like to capture the total electrical power consumption by measuring the battery energy drop from start to end of each trial. Unfortunately, because Minitaur's direct drive actuators have high specific power but low specific torque \cite{kenneally_design_2016}, and because the pronk is a highly energetic (not efficient \cite{McMahon_1985}) gait, rapid Joule heating limits the safe trial duration to a distance whose total energy expenditure is insufficient to be reliably measured using readily available commercial technology, e.g.,  as in \cite{Roberts_Koditschek_2021}. Instead, we compute an approximation of the total cost of transport as the sum of average  mechanical output power expended in the fore-aft direction added to the electrical power lost in Joule heating.

As a rough approximation to the average mechanical power consumed in the accumulation of fore-aft distance, we treat each leg as a ``single-spoked wheel" whose radius is given by the average limb displacement during stance, $\bar{\chi}$, and whose horizontal work is achieved by the average torque imposed upon the ``wheel axle" by the two hip motors for each limb. The average torque ($\bar\tau$) is computed by averaging the current drawn by the two hip motors (per time step) for each limb $i$, a $n \times 1$ vector denoted as $\bar{\textbf{I}}_{avg_i}$, then summing the average currents across all limbs (per time step), and multiplying the resulting vector by the motor torque constant. The resulting estimate takes the form
\begin{equation} \label{eq1}
\begin{split}
\bar{\textbf{P}}_{mechanical} &= 2 \bar\tau \bar{\dot{\Theta}}  \\
 & = 2 K_{\tau} (\bar{\textbf{I}}_{avg_{LF}}+\bar{\textbf{I}}_{avg_{LB}}+\bar{\textbf{I}}_{avg_{RF}}+\bar{\textbf{I}}_{avg_{RB}}) \bar{\dot{\Theta}} \\
& = 2 K_{\tau} \bar{\textbf{I}}_{avg} \bar{\dot{\Theta}} 
\end{split}
\end{equation}
where $\bar{\dot{\Theta}}  = \frac{\bar{\textbf{v}}}{\bar\chi}$ is the fore-aft speed of the ``single-spoked wheel" during stance, $K_{\tau}$ is the motor torque constant, and $\bar{I}_{avg_{i}}$ is the average motor current for limb $i$.

% write out what I^2 total is
In contrast, the average electrical power lost by the robot to Joule heating is 
\begin{equation} \label{eq:pwr_heat}
\begin{split}
    \bar{\textbf{P}}_{heat} &= (\bar{\textbf{P}}_{heat_{m1}} + \bar{\textbf{P}}_{heat_{m2}})/2 \\
    &= (\bar{\textbf{I}}_{m1}^2 \text{R} + \bar{\textbf{I}}_{m2}^2 \text{R})/2
\end{split}
\end{equation}
where $\bar{\textbf{I}}_{m1}$ and $\bar{\textbf{I}}_{m2}$ are the time averaged current vectors for motor one and motor two, respectively, and R is the motor's internal resistance.

We now add these two components of output power to obtain the numerator of (\ref{eqn:specific})
\begin{equation}
    \bar{\textbf{P}} = \bar{\textbf{P}}_{mechanical} + \bar{\textbf{P}}_{heat}
\end{equation}

\section*{ACKNOWLEDGMENT}
We thank Dr. Wei-Hsi Chen and Dr. Sonia Roberts for a number of very helpful discussions.

This work was supported in part by an NSF Graduate Research Fellowship held by the first author and in part by the Office of Naval Research grant N00014-16-1-2817, awarded as a Vannevar Bush Faculty Fellowship held by the second author, sponsored by the Basic Research Office of the Assistant Secretary of Defense for Research and Engineering.

% end of acknowledgment
%%%%%%%%%%%%%%%%%%%%%%%%%%%%%%%%%%%%%%%%%%%%%%%%%%%%%%%%%%%%%%%%%%%%%%%%%%%%%%%%

\printbibliography
% end of references

\end{document}